\documentclass[letterpaper]{article}

\usepackage{aaai2026}
\usepackage{times}
\usepackage{helvet}
\usepackage{booktabs}
\usepackage{courier}
\usepackage[hyphens]{url}
\usepackage{graphicx}

\usepackage{natbib}
\usepackage{caption}

\usepackage{textgreek}

\usepackage{algorithm}
\usepackage{algorithmic}

\usepackage{newfloat}
\usepackage{listings}

\DeclareCaptionStyle{ruled}{
    labelfont=normalfont, we
    labelsep=colon,
    strut=off
}

\floatstyle{ruled}
\newfloat{listing}{tb}{lst}{}
\floatname{listing}{Listing}

\usepackage{amsmath}

\nocopyright

\title{Design Theater: Evaluating the Gap Between User-Facing Design Reasoning and Implementation in Generative UI Tools}

\author{
    Kashif Imteyaz\textsuperscript{\rm 1}\thanks{Corresponding author.},
    Kaif Imteyaz\textsuperscript{\rm 2},
    Nakul Rajpal\textsuperscript{\rm 1},
    Kaif Shaikh\textsuperscript{\rm 3},
    Michael Muller, \textsuperscript{\rm 4},
    Saiph Savage\textsuperscript{\rm 1,5}
}

\affiliations{
    \textsuperscript{\rm 1}Northeastern University, United States\\
    \textsuperscript{\rm 2}Jamia Hamdard, India\\
    \textsuperscript{\rm 3}Saarland University, Germany\\
    \textsuperscript{\rm 4}Independent Researcher, United States\\
    \textsuperscript{\rm 5}Universidad Nacional Autonoma de Mexico (UNAM), Mexico\\
    imteyaz.k@northeastern.edu
}

\begin{document}

\maketitle

\begin{abstract}
Generative UI tools promise to democratize UI design by turning natural language descriptions into complete interfaces. Alongside the interface, these tools generate user-facing design rationales that explain their layout, accessibility, and design choices. However, it remains unclear whether these stated rationales are actually reflected in the interfaces they produce. We call this disconnect ``Design Theater'': plausible and confident design rationales that have little relationship to the actual implementation. To study this phenomenon, we introduce a benchmark and three metrics for measuring Design Theater. The benchmark includes 24 UI generation tasks spanning structural, styling, and functional design requirements. Using this benchmark, we evaluate 120 interfaces created by five generative UI tools. On average, roughly 25\% of user-facing design rationales are not implemented in the generated interface, and the implementation failure increases to 34\% for functional requirements. Tools recognize roughly half of the UX principles embedded in prompts (mean = 0.54), with four of five tools implementing 6\% or fewer functional principles. We also measure interface similarity across tools and find convergence in visual appearance and layout organization, with greater variation in color choices. Overall, we contribute: 1) the concept of Design Theater; 2) a benchmark with metrics for assessing whether the stated reasoning of generative UI tools is reflected in their implementations; 3) and findings from a systematic evaluation of these tools. We discuss what these findings mean for the design and evaluation of generative UI tools.

\end{abstract}


\section{Introduction}
Generative UI tools are increasingly positioned as a way to lower barriers to interface design by allowing users to generate interface prototypes from natural-language prompts \cite{takaffoli2024generative, chen2025genui, zhou2024exploring}. Recent work describes such tools as valuable not only for UX designers, but also for adjacent roles such as developers and product managers \cite{chen2025genui}. Generative UI tools such as Vercel v0, Claude Artifacts, ChatGPT Canvas, Firebase Studio, and Bolt support prompt-based workflows in which users describe an interface or application idea and receive generated code, previews, or deployable prototypes \cite{vercel2023v0, anthropic2026artifacts, openai2024canvas, google2026firebasestudio, stackblitz2026bolt}.

Beyond generating interfaces, these tools often narrate their design process through what we call \textit{user-facing design reasoning}: explanations of what they intend to create, why particular design decisions are appropriate, and how the output addresses layout, accessibility, color, interaction, or broader usability principles \cite{nielsen1994usability}. For first-time creators, product managers, engineers, or other non-design stakeholders, such reasoning can function as a signal of design competence, making generated interfaces appear more deliberate, principled, and trustworthy \cite{liao2022designing, sun2026seeing}. As AI-generated interfaces move from experimentation into organizational design and development workflows, this reasoning may become part of how teams explain, review, and justify design decisions \cite{son2026clearfairy, takaffoli2024generative}. As a result, users may treat the tool's reasoning as evidence that usability, accessibility, and interaction-design concerns were properly addressed, even when they lack the expertise to verify \cite{kaur2020interpreting, sun2026seeing}. For example, when a generative UI tool states that it will create ``a responsive three-column grid with accessible contrast ratios and keyboard-navigable tabs'', it creates the impression of careful, principle-driven design. Yet it remains unclear whether this reasoning is actually reflected in the generated interface. This is especially concerning as design work is increasingly delegated to AI tools and reviewed by product managers, engineers, or other stakeholders who may not have formal design training. Without methods for evaluating whether generative UI tools follow through on their user-facing design reasoning, teams may over-trust interfaces that appear professionally designed while failing to implement critical usability, accessibility, or responsiveness requirements \cite{liao2022designing}.

To better understand and study this phenomenon, we introduce the concept of \textit{Design Theater}: the production of user-facing design reasoning that bears little relationship to the actual design implementation. We define it by its effect on the reader. A rationale written in the language of professional design reads as evidence of deliberate, principle-driven decision-making, whether or not those commitments are realized in the artifact. Persuasive reasoning invites less scrutiny of the generated artifact, allowing overreliance to emerge before verification occurs \cite{sun2026seeing}.

However, defining Design Theater is not enough. We also need benchmarks and metrics that can measure how often generative UI tools produce mismatches between their user-facing design reasoning and actual implementations. Quantifying Design Theater allows researchers and practitioners to move beyond anecdotal examples, compare tools systematically, and identify when interfaces appear trustworthy despite failing to implement key usability, accessibility, or interaction-design requirements.

To address this need, this paper presents a benchmark and three metrics for measuring Design Theater in generative UI tools. The benchmark evaluates reasoning-to-implementation alignment and design homogenization across 120 UI designs generated by five state-of-the-art UI agent tools across 24 design tasks. These tasks span three tiers of design tasks: structural tasks, including layout and information architecture; styling tasks, including color, typography, and spacing; and functional tasks, including interactivity and user flows. We pair the benchmark with three metrics: Thinking Fidelity Score (TFS), Principle Adherence Score (PAS), and Design Homogeneity Index (DHI). Together, these metrics quantify whether tools’ user-facing design reasoning is reflected in their generated interfaces and whether different tools produce similar designs for the same prompt. Using this benchmark and set of metrics, we study the following research questions:

\begin{enumerate}
    \item To what extent is the user-facing design reasoning produced by generative UI tools reflected in the interfaces they implement?
    \item To what extent do generative UI tools recognize and implement UX principles that are implicitly embedded in natural-language prompts (tasks)?
    \item How much do generated interfaces vary in their visual appearance, color schemes, and layouts when different generative UI tools are given the same task?
\end{enumerate}

Our findings suggest that, on average, roughly 25\% of user-facing design reasoning is not implemented in the generated interface, with implementation failures increasing to 34\% for functional requirements. Tools recognize roughly half of the UX principles implicitly embedded in prompts (mean PAS = 0.54), with four of the five tools implementing 6\% or fewer functional principles. We also find convergence across tools: when given the same prompt, generated interfaces showed narrow pairwise differences in visual appearance and layout organization, while color choices varied more widely. Based on these findings, we discuss the implications for how generative UI tools are evaluated and for the stakeholders who increasingly review their output.

Our paper makes the following contributions:

\begin{itemize}
    \item \textbf{Conceptualization of Design Theater.} We introduce the concept of \textit{Design Theater}, describing how generative UI tools produce plausible design rationales that may not correspond to their implemented interfaces.


    \item \textbf{A benchmark and metrics for evaluating Design Theater.} We introduce a benchmark with 24 UI design tasks and three complementary metrics, TFS, PAS, and DHI, for evaluating reasoning-to-implementation alignment, implicit UX principle adherence, and cross-tool design homogenization.

    \item \textbf{Empirical evaluation of generative UI tools.} Through a systematic study of 120 interfaces generated by five UI tools across 24 design tasks, we show that misalignment between user-facing reasoning and implementation is widespread and concentrates in functional design. We find that designs converge in visual appearance and layout when tools are given the same prompt.
\end{itemize}

\section{Related Work}

\subsection{Vibe Coding and AI-Assisted Co-Creation}



Vibe coding is an emerging practice in which developers build software by describing intent in natural language rather than writing code directly~\cite{li2026vibe}. Its recent rise aligns with coding tools such as OpenAI Codex, Claude Code, Cursor, and GitHub Copilot~\cite{openai2025codex, cursor2024, anthropic2025claudecode, github2021copilot}, which promise to reduce barriers to software development. These coding agents can produce code that compiles and runs, though prior work suggests that generated code matches users' intended outputs only at moderate rates~\cite{yeticstiren2023evaluating}.

Researchers have therefore begun examining how users interact with coding agents. A common theme is the redistribution of effort from writing code to evaluating AI output and managing conversational context~\cite{sarkar2025vibecodingprogrammingconversation}. In this workflow, users may skip quality assurance steps, reducing the reliability of the final output~\cite{fawzy2025vibecodingpracticemotivations}. These omissions can carry real consequences: studies have found that 40\% of GitHub Copilot-generated code contained security weaknesses~\cite{pearce2025asleep}, and an analysis of production repositories found similar issues in 30\% of Python and 24\% of JavaScript snippets~\cite{yujia2025SecurityWeakness}.
 
These concerns are not limited to code correctness. AI-assisted creation also raises questions about output diversity~\cite{doshi2024generative}. In a large-scale writing experiment, access to a generative AI tool improved individual output quality but reduced collective output diversity across participants~\cite{doshi2024generative}. Similar homogenization effects have been identified in design tasks~\cite{wadinambiarachchi2024effects} and creative ideation~\cite{anderson2024homogenization}. Related work also shows that generative systems can default to Western cultural conventions~\cite{dhruv2025WesternStyle}, a concern that is relevant for interface design, where layout, color, typography, and interaction patterns are culturally situated.

Concerns about homogenization also build on a longer history of web design convergence. \citet{goree2021investigating} found that web designs became more similar after 2007, with layout distance alone dropping over 30\%, driven largely by the adoption of shared libraries and frameworks like Bootstrap. Our work asks whether generative UI tools extend this trajectory while obscuring it, layering rationales of deliberate design choice over outputs that may default to a narrow band of conventions.

Beyond generating outputs, these systems increasingly narrate their process through user-facing reasoning 
that explain why particular choices were made. Prior work suggests that users may scan or rely on these traces instead of evaluating outputs directly~\cite{fawzy2025vibecodingpracticemotivations}. What remains unexamined is whether such traces accurately reflect what was built, and whether users can distinguish competent design from a persuasive description of competence.


\subsection{From Design Templates to Generative UI Tools}




Early low-code and template-based interface tools lowered the barrier to creating interactive systems by allowing users to assemble interfaces from predefined components rather than writing code from scratch. While simple and effective, these tools constrained users to the structures and interaction patterns provided by the template system. Subsequent research moved beyond templates toward machine-learning systems for translating interface images or sketches into code, detecting UI components, and recovering interface structure from screenshots~\cite{beltramelli2018pix2code, jain2019sketch2code, wu2021screenparsing, wu2023never-ending}. These systems advanced computational interface understanding, but often remained tied to specific input modalities, datasets, or interaction domains.

The integration of large language models further expanded design-generation workflows beyond fixed templates and task-specific models. LLM-powered systems now support a range of creative design tasks, including text-to-image product design~\cite{tao2025designweaver}, animation design~\cite{tseng2024keyframer, liu2024logomotion}, and interactive story generation~\cite{chung2022talebrush}, and the use of generated images as probes in co-design~\cite{imteyaz2026co}.

More recently, generative UI tools have extended this trajectory by producing interface artifacts from natural-language prompts. Tools such as Vercel v0~\cite{vercel_v0_2026}, Lovable~\cite{lovable2025}, and Replit~\cite{replit2026} support prompt-based workflows for generating interface code, previews, and deployable prototypes. Many also produce user-facing reasoning traces that explain the design choices behind the generated interface. This shift has prompted renewed attention to computational approaches for UI understanding, generation, and evaluation~\cite{jiang2023evaluser, peng2026human}, including work arguing that interfaces must increasingly be designed not only for human users but also for AI agents acting on users' behalf~\cite{peng2026human}.

Although these tools can produce visually plausible and functional artifacts, many evaluations of AI code-generation systems still emphasize functional or code-level correctness~\cite{chen2021evaluatinglargelanguagemodels, yeticstiren2023evaluating}. Design quality, however, is not reducible to functional correctness~\cite{stone2005user}. A UI that renders without errors can still exhibit poor information architecture, incoherent visual hierarchy, inaccessible interaction patterns, or cultural and aesthetic biases embedded in training data~\cite{dhruv2025WesternStyle}. Current generative UI tools offer limited support for verifying whether these qualities are present in the generated output, particularly when the tool's own reasoning traces suggest they have been addressed.

\subsection{LLM Reasoning Traces and User Interpretation}


Chain-of-thought (CoT) prompting instructs language models to generate intermediate reasoning steps before producing a final answer, improving performance on arithmetic, common-sense, and symbolic reasoning tasks~\cite{wei2022chain, kojima2022large, wang2022self}. Subsequent work has treated these generated rationales as both a performance enhancer and a possible window into model behavior.

However, a separate line of work questions whether these traces faithfully reflect the model's internal reasoning. When biased features are introduced into prompts, model answers can shift while the accompanying rationales fail to account for those shifts, producing plausible but unfaithful explanations~\cite{turpin2023language}. Other work finds that early reasoning steps can be modified without changing final outputs, further suggesting that generated rationales may function as post-hoc reconstructions rather than faithful process logs~\cite{lanham2023measuring, ji2023survey}.

The faithfulness of generated rationales is not only a model-centric concern; it also affects how users interpret and rely on AI systems. Empirical work shows that users treat reasoning traces as trust-calibration tools, using them to interpret code explanations and decide whether to accept or revise AI outputs~\cite{sun2026seeing}. The presentation of rationales alone can shape trust in AI decisions, even when the underlying model behavior remains unchanged~\cite{sun2026seeing, bertrand2022cognitive}.

Most work on reasoning-trace faithfulness examines tasks with externally checkable answers, such as arithmetic, fact verification, and symbolic reasoning~\cite{wei2022chain, turpin2023language, chen2025reasoning, bertrand2022cognitive}. Design reasoning presents a different challenge. UI outputs are visual and interactive artifacts whose quality depends on contextual, multi-dimensional, and often subjective criteria, including usability, accessibility, visual hierarchy, and interaction flow. Unlike this literature, which asks whether rationales reflect the model's internal reasoning process, we ask a downstream question: whether the rationale is consistent with the artifact the user actually receives. For non-expert stakeholders, who often read these rationales as design documentation, the key issue is whether the rationale accurately describes the resulting artifact. Whether this relationship holds in open-ended creative tasks such as UI generation remains unexplored.

\subsection{Automated Design Evaluation and UX Metrics}


Usability evaluation has a longer history than its current automated forms suggest. Before heuristic methods became central in HCI, human factors research developed standardized instruments for evaluating workload, usability, situation awareness, and readability in safety-critical domains~\cite{hart2006nasa, brooke1996sus, wickens2021engineering}. Heuristic evaluation entered this landscape as a deliberately lightweight alternative~\cite{nielsen1994usability}, valued for making usability problems inspectable without requiring full user studies. We draw on heuristic and guideline-based evaluation pragmatically: these frameworks do not exhaust design quality, but they provide inspectable commitments that can be checked against generated artifacts. Since design quality is situated and shaped by context, values, and power relations~\cite{haraway2013situated, costanza2020design}, our benchmark evaluates whether tools implement the commitments they themselves make, without claiming that any checklist can fully determine design appropriateness. Contextual Inquiry~\cite{karen2017contextual} and Scandinavian Participatory Design~\cite{ehn2017scandinavian} offer a complementary tradition, grounding design assessment in real work contexts and the expertise of affected users rather than expert inspection alone. This tradition reminds us that heuristic and guideline-based evaluation can be useful without fully capturing situated design quality.

Alongside heuristic methods, design communities developed more implementable specifications, including platform interface guidelines~\cite{apple2023hig} and the Web Content Accessibility Guidelines~\cite{WCAG22}. Automated evaluation tools later operationalized some of these specifications and perceptual principles. Axe-core and Lighthouse evaluate accessibility and performance against established web standards~\cite{deque2024axecore, google2024lighthouse}, while Aalto Interface Metrics computes perceptual measures such as visual clutter and learnability~\cite{antti2018AIM}. More recent AI-driven methods provide design feedback on layout, color, and typography~\cite{lee2020GUIcomp}, predict visual attention on interfaces~\cite{Jiang2023UEyes}, or use LLMs as design critics grounded in heuristic and guideline-based datasets~\cite{duan2024UICrit, jiang2023evaluser}.

Across these approaches, evaluation criteria are typically grounded in external specifications such as Nielsen's heuristics, WCAG, platform guidelines, perceptual models, or human-factors instruments. However, they generally do not examine whether a tool's \textit{own stated design reasoning} is reflected in its generated output. A tool may provide a rationale explaining that it prioritizes accessibility, describe its color choices in terms of WCAG contrast ratios, and narrate a deliberate information hierarchy, yet produce an interface that contradicts those commitments. Our work addresses this gap by treating usability, accessibility, and interaction-design principles as rationales that can be verified against generated artifacts, rather than assurances to be accepted.


\section{Methodology}
Our methodology is designed to quantify \textit{Design Theater}: mismatches between the user-facing design reasoning produced by generative UI tools and the interfaces they actually implement. To do so, we introduce a benchmark composed of 24 UI generation tasks and three evaluation metrics: Thinking Fidelity Score (TFS), Principle Adherence Score (PAS), and Design Homogeneity Index (DHI). Together, the benchmark and metrics allow us to assess whether generative UI tools follow through on their stated design reasoning, recognize UX principles embedded in natural-language prompts, and converge toward similar interface designs, revealing potential design homogenization.


\subsection{Generative UI Tools}
We evaluate five widely used generative UI tools: ChatGPT, Claude, Firebase Studio, Vercel v0, and Bolt \cite{vercel2023v0, anthropic2026artifacts, openai2024canvas, google2026firebasestudio, stackblitz2026bolt}. We selected tools that: (1) accept natural-language descriptions as input; (2) generate HTML, CSS, JavaScript, or equivalent frontend code; and (3) produce user-facing reasoning traces that explain their design decisions.

For each tool, we used its default user-facing configuration. We did not modify system-level or developer-level prompts, and we left each tool’s default model settings unchanged. This setup reflects how non-expert creators are likely to encounter and use these tools in practice. 

\subsection{Benchmark Design}
We introduce a benchmark of 24 UI generation tasks designed to evaluate reasoning-to-implementation alignment in generative UI tools. In the benchmark, each task consists of a natural-language prompt that asks a generative UI tool to create a complete interface for a particular scenario. The prompts are designed to implicitly embed UX principles through user needs, contextual constraints, audience characteristics, and usage scenarios, rather than explicitly instructing the tool to implement specific design principles. This setup allows us to evaluate whether generative UI tools can recognize and operationalize fundamental UX requirements that are implied by the prompt, reflecting how novice designers and non-expert users are likely to request interfaces in practice \cite{chen2025genui, raees2026understanding, zamfirescu2023johnny}.

Our benchmark is organized into three tiers of design tasks that reflect core dimensions of UX design: structural organization, visual presentation, and functional interaction \cite{rosenfeld2015information, Jiang2023UEyes, nielsen1994usability}.
Each tier 
foregrounds a different kind of design work, allowing us to examine where Design Theater might appear in the generated interfaces. Structural tasks help us to examine how generative UI tools organize information, styling tasks examine how tools make visual design decisions, and functional tasks examine how tools implement interactive behavior. This organization allows us to study whether reasoning-to-implementation gaps emerge differently across what an interface: 1) contains; 2) how it looks; and 3) how it behaves.

\subsubsection{Tier 1: Structural Tasks}
Structural tasks ask generative UI tools to create interfaces where the central challenge is how information is structured \cite{rosenfeld2015information}. These tasks focus on whether tools can organize information in ways that help users understand what content exists, how different pieces of content relate to each other, where to find relevant information, and what actions to take next. We use these tasks to study whether tools can translate reasoning about information structure into concrete interface decisions, such as grouping related content, ordering sections meaningfully, creating navigation paths, routing different user groups, and surfacing high-priority information. The tasks progress from simple single-audience interfaces, such as a local business website, to more complex civic or institutional systems where multiple user groups must locate different types of information.

\subsubsection{Tier 2: Styling Tasks}
Styling tasks ask generative UI tools to create interfaces where the main design challenge is visual communication \cite{Jiang2023UEyes}. These tasks help us study whether tools can translate visual design reasoning into concrete styling decisions, such as color choices, typography, spacing, visual hierarchy, brand expression, readability, accessibility, and emotional tone. The tasks progress from straightforward branding scenarios to more complex interfaces for users with constrained devices, varying literacy levels, or heightened emotional stress.

\subsubsection{Tier 3: Functional Tasks}
Functional tasks ask generative UI tools to create interfaces where the central challenge is how the interface behaves during user interaction \cite{nielsen1994usability}. These tasks focus on whether tools can implement interactive behavior that supports users as they move through a workflow, encounter errors, change input states, navigate between roles, or use accessibility features. We use these tasks to study whether tools can translate interaction-design reasoning into working interface behavior. The tasks progress from simple workflows, such as a reservation flow where users select a date, choose a time, enter their information, and confirm a booking, to more complex interactive systems that must serve large and diverse user populations.

Within each tier, tasks increase in complexity along four dimensions. First, they vary in the number of user audiences, from interfaces designed for one group to interfaces that must serve multiple groups, such as residents, administrators, and visitors. Second, they vary in the number of competing design requirements, such as balancing simplicity, accessibility, branding, and detailed information. Third, they vary in the depth of domain-specific constraints, such as bilingual content, regulatory requirements, limited device access, or low literacy. Fourth, they vary in the severity of consequences if the design fails, ranging from confusion in a low-stakes website to missed emergency, health, or financial information in a high-stakes interface. This progression allows us to examine whether Design Theater becomes pronounced as design scenarios become more complex.

\subsection{Prompting and Generative UI Output Collection}
We administered the 24 benchmark tasks, eight per tier, to each of the five generative UI tools in our study, producing 120 generated interfaces in total ($5$ tools $\times$ $24$ tasks). For each tool-task combination, the prompt included the same fixed implementation instruction: the tool had to generate the interface using only HTML, CSS, and JavaScript, without external frameworks, libraries, or web searches. We kept this instruction identical across all tools and tasks so that differences in the generated interfaces would reflect each tool's interpretation of the task prompt, user-facing design reasoning, and implementation choices, rather than differences in external resources or selected libraries. For each output, we captured the generated code, the rendered interface, and the full user-facing reasoning trace, including design commentary, explanations, and any rationale produced before, during, or alongside code generation.




\subsection{Metric 1: Thinking Fidelity Score (TFS)}
TFS measures the extent to which a generative UI tool’s user-facing design reasoning is reflected in the interface it generates. TFS operationalizes Design Theater by quantifying how much of the concrete design reasoning articulated by the tool is fully, partially, or not implemented in the final interface. In order to calculate this metric, we conduct:

\paragraph{1) Reasoning Element Extraction.}
For each reasoning trace, we extract every concrete and verifiable design reasoning element. A reasoning element must be specific enough to evaluate against the generated interface or code. For example, statements such as ``I will use a three-column grid,'' ``I will include keyboard-navigable tabs,'' or ``I will add a dark mode toggle'' are included. Vague reasoning, such as ``I will make the design user-friendly,'' is excluded because it cannot be reliably verified against the final interface.

\paragraph{2) Reasoning Verification.}
Two independent evaluators compare each extracted element of user-facing design reasoning against the generated interface. Each reasoning element is classified as \textit{Fully Implemented} (1.0), \textit{Partially Implemented} (0.5), or \textit{Not Implemented} (0.0), depending on the extent to which the stated design reasoning is realized in the implemented interface. Each reasoning element is scored against the generated interface rather than the task prompt. An element is counted as Not Implemented when the interface does not contain what the reasoning described, including cases where the tool implemented a different approach. Whether the interface meets the prompt's requirements is measured separately by PAS.

\paragraph{3) TFS Metric Computation.}
\begin{equation}
\text{TFS} = \frac{1.0 \cdot N_{\text{full}} + 0.5 \cdot N_{\text{partial}}}{N_{\text{total}}}
\end{equation}

A lower TFS indicates greater Design Theater because the tool's user-facing reasoning is less faithfully reflected in the generated interface. TFS measures the size of this gap, not
its cause.

\subsection{Metric 2: Principle Adherence Score (PAS)}
PAS measures whether generative UI tools recognize and implement UX principles that are implicitly embedded in the prompt. Whereas TFS evaluates whether tools follow through on their own stated reasoning, PAS evaluates whether tools implement the underlying UX principles required by each prompt within a task (each task has two UX principles). To avoid defining these principles post hoc, we use a framework-first approach: we select UX principles from established HCI and design frameworks before evaluation, and then design each prompt to naturally embed those principles through user needs and contextual constraints. Structural tasks draw from information architecture principles \cite{rosenfeld2015information}, styling tasks draw from visual perception and WCAG-related accessibility principles \cite{Jiang2023UEyes, w3c2024wcag22}, and functional tasks draw from usability heuristics \cite{nielsen1994usability}. To calculate PAS, we use:

\paragraph{1) Scoring Protocol.}
Each prompt (task) contains two target UX principles. For each generated interface, two independent evaluators score whether each principle was implemented: 1 if implemented and 0 if not implemented.

\paragraph{2) PAS Metric Computation.}
\begin{equation}
\text{PAS} = \frac{\sum_{i=1}^{2} P_i}{2}
\end{equation}
where $P_i \in \{0,1\}$ indicates whether principle $i$ is implemented.


For both TFS and PAS, evaluators first completed their ratings independently. Disagreements were then resolved through discussion, and any remaining unresolved cases were adjudicated by a third evaluator before final metric computation \cite{o2020intercoder, mcdonald2019reliability}. Inter-rater reliability on the independent ratings was substantial for TFS (linearly weighted Cohen's κ = 0.70; per-tool range 0.53–0.85) and almost perfect for PAS (κ = 0.90; per-tool range 0.75–1.00). Weighted κ is used for TFS because the classification is ordinal, and disagreements were almost entirely adjacent.

\subsection{Metric 3: Design Homogeneity Index (DHI)}
DHI measures design homogenization by examining how similar the generated interfaces are across tools, especially when those tools are given the same task prompt. Because UI similarity is multidimensional \cite{Jiang2023UEyes}, we do not collapse DHI into a single aggregate score. Instead, we report DHI as an index composed of three sub-measures that capture different aspects of interface similarity: overall visual appearance, color usage, and layout structure.

To calculate DHI, we first render all 120 generated interfaces as standardized 1200 $\times$ 1200 PNG screenshots. For each prompt, we compare the five interfaces generated by the five tools. This yields ten pairwise tool comparisons per prompt for each DHI sub-measure.

\paragraph{DHI-Visual Similarity.}
We encode each screenshot using UIClip and compute pairwise visual similarity across tools for the same prompt \cite{wu2024uiclip}. This sub-measure captures the overall visual appearance of the interface, including stylistic patterns, component aesthetics, spacing tendencies, and the general visual impression produced by the design.

\paragraph{DHI-Color Similarity.}
We represent each screenshot as a color histogram in CIELCh color space and compute pairwise color distance using Earth Mover's Distance \cite{wyszecki2000color, rubner2000earth}. This submeasure specifically captures how similar the interfaces are in their use of colors, including palettes, color balance, saturation, and color distribution throughout the interface.

\paragraph{DHI-Layout Similarity.}
We parse each screenshot using OmniParser v2.0  to detect UI elements, and then compute structural similarity using tree edit distance \cite{lu2024omniparser}. This submeasure captures how similarly interface components are organized and positioned, including section ordering, navigation placement, grouping of elements, and overall page structure.

We report each DHI sub-measure separately because visual appearance, color usage, and layout structure can capture distinct dimensions of design homogenization. This allows us to identify not only whether generative UI tools converge, but also where that convergence occurs: in stylistic appearance, color choices, or interface layout organization.



\section{Results}

\paragraph{Metric 1: Thinking Fidelity Score (TFS)} Using our TFS metric, we measured the proportion of stated design reasoning that each generative UI tool actually implemented across its generated interfaces. 

\begin{figure}[t]
\centering
\includegraphics[width=\columnwidth]{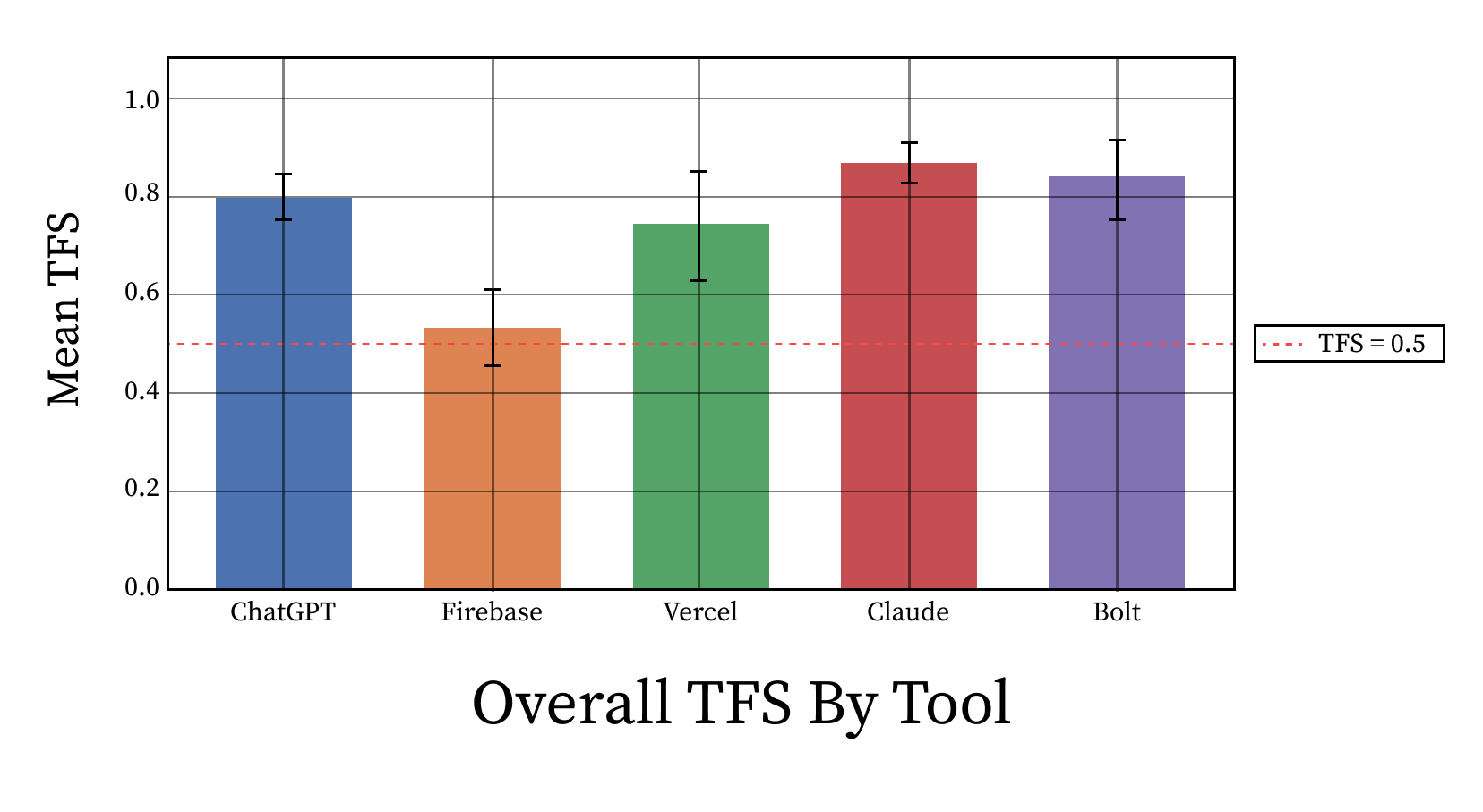}
\caption{Overall Thinking Fidelity Score by Tool}
\label{fig:TFS-tierxTool}
\end{figure}

Figure~\ref{fig:TFS-tierxTool} shows the average TFS scores for each tool across its generated interfaces, while Table~\ref{tab:tfs_results} in its second column reports the overall TFS results. No tool consistently delivered on its design reasoning, and the cross-tool mean TFS score was 0.75, indicating that roughly one in four stated design rationales did not fully appear in the generated interface. Four tools (Claude, Bolt, ChatGPT, and Vercel v0) clustered between 0.74 and 0.87, suggesting descriptively similar overall performance, with Claude achieving the highest alignment at 0.87. Firebase scored lowest, with a mean TFS of 0.53, below the range observed for the other four tools. 

\begin{table}[t]
\centering
\small
\setlength{\tabcolsep}{4pt}

\resizebox{\columnwidth}{!}{%
\begin{tabular}{lcccc}
\toprule
\textbf{Tool} & \textbf{Overall} & \textbf{Tier 1} & \textbf{Tier 2} & \textbf{Tier 3} \\
              &                  & Structural      & Styling         & Functional \\
\midrule
Claude     & \textbf{0.87} [0.82, 0.91] & 0.83 [0.74, 0.91] & 0.93 [0.89, 0.96] & \textbf{0.85} [0.78, 0.91] \\
Bolt       & 0.84 [0.75, 0.91] & 0.83 [0.70, 0.95] & \textbf{0.95} [0.90, 0.99] & 0.73 [0.55, 0.89] \\
ChatGPT    & 0.80 [0.75, 0.84] & \textbf{0.85} [0.75, 0.94] & 0.78 [0.72, 0.85] & 0.75 [0.69, 0.81] \\
Vercel v0  & 0.74 [0.63, 0.84] & 0.83 [0.70, 0.94] & 0.83 [0.66, 0.96] & 0.56 [0.36, 0.77] \\
Firebase   & 0.53 [0.45, 0.61] & 0.62 [0.54, 0.69] & 0.55 [0.38, 0.70] & 0.43 [0.34, 0.56] \\
\midrule
\textit{Across tools} & \textit{0.75} & \textit{0.79 } & \textit{0.81 } & \textit{0.66 } \\
\bottomrule
\end{tabular}%
}
\caption{Mean Thinking Fidelity Scores (TFS) by tool and tier. Best score in each column shown in bold. }
\label{tab:tfs_results}
\end{table}

\begin{figure}[t]
\centering
\includegraphics[width=\columnwidth]{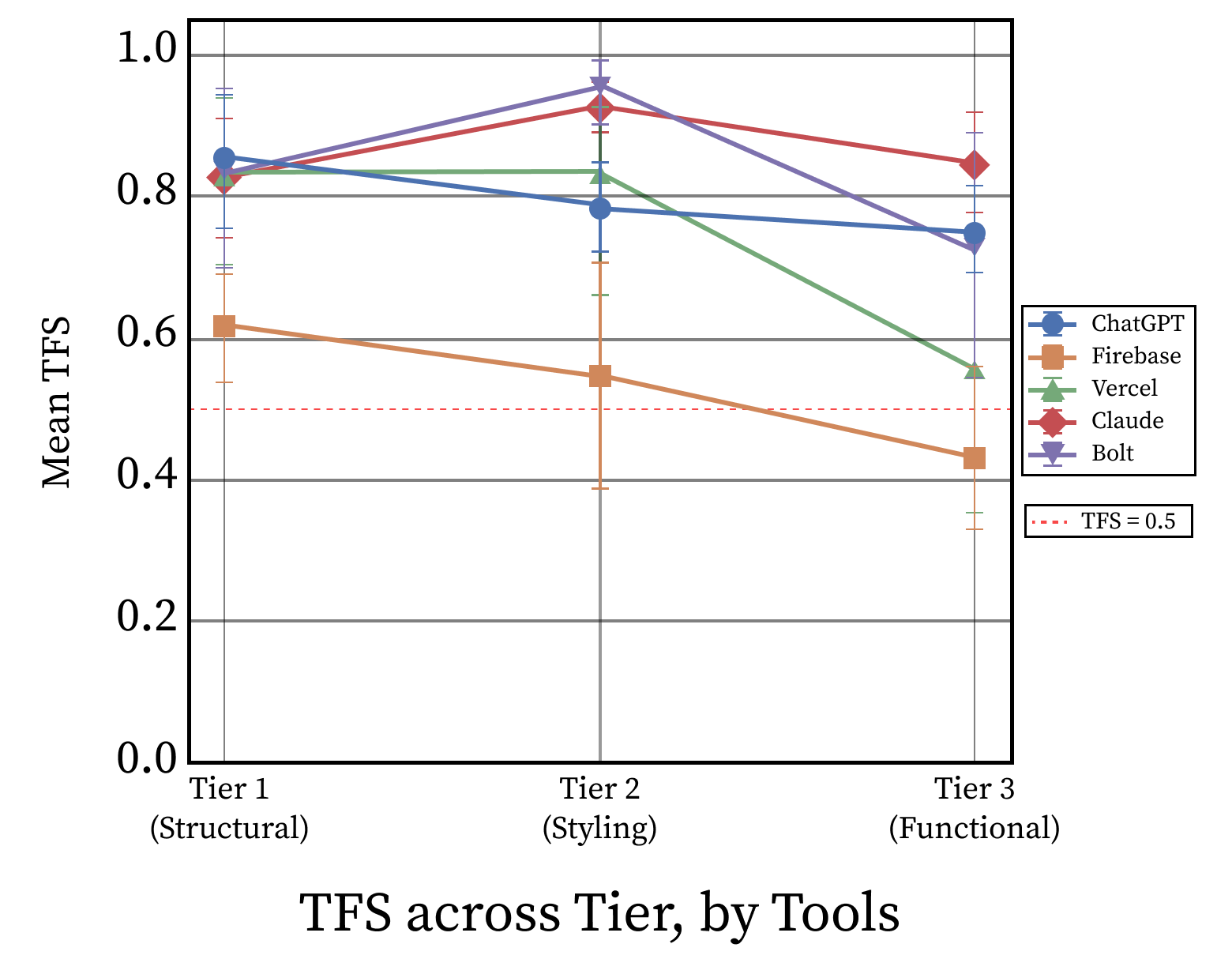}
\caption{Thinking Fidelity Score across Tier, by Tools}
\label{fig:TFSTierandTool}
\end{figure}

Figure~\ref{fig:TFSTierandTool} and Table~\ref{tab:tfs_results} show that reasoning-to-implementation fidelity varied across both tools and task tiers. Functional tasks showed the weakest alignment across tools, with a cross-tool mean TFS of 0.66 compared with 0.79 for structural tasks and 0.81 for styling tasks. However, this decline was not uniform across tools. Claude maintained relatively stable fidelity across tiers, scoring 0.83 on structural tasks, 0.93 on styling tasks, and 0.85 on functional tasks. Bolt peaked on styling tasks before declining on functional tasks, while ChatGPT showed a gradual decline from structural to styling to functional tasks. Vercel v0 performed similarly on structural and styling tasks before dropping sharply on functional tasks, and Firebase remained the lowest-performing tool across all tiers. Together, these results suggest that Design Theater is not distributed evenly across task types: for some tools, reasoning-to-implementation gaps are broadly distributed, while for others they are concentrated especially in functional implementation.


Table~\ref{tab:tfs_results} also shows that tools differed not only in average fidelity but also in the uncertainty around those estimates. Claude and ChatGPT had the narrowest overall bootstrap confidence intervals, with Claude ranging from 0.82 to 0.91 and ChatGPT from 0.75 to 0.84. In contrast, Vercel v0 and Bolt showed wider intervals, especially for Tier 3 functional tasks. For example, Vercel v0’s Tier 3 mean was 0.56, with a confidence interval from 0.36 to 0.77. This suggests greater variability in how reliably Vercel v0 translated functional design claims into implementation across tasks. In some cases, the tool appeared to follow through on its stated reasoning, while in others, much of the promised functionality was absent or incomplete. These results indicate that mean TFS alone can obscure differences in reliability: a tool with moderate but stable fidelity presents a different risk profile than one whose performance appears more variable across tasks.

Overall, from the perspective of a stakeholder reading a tool's reasoning trace, the practical question is not which tool to trust most, but how any tool's reasoning should be read as a description of what was actually built.

\paragraph{Metric 2: Principle Adherence Score (PAS)}
\begin{figure*}[t]
\centering
\includegraphics[width=\textwidth]{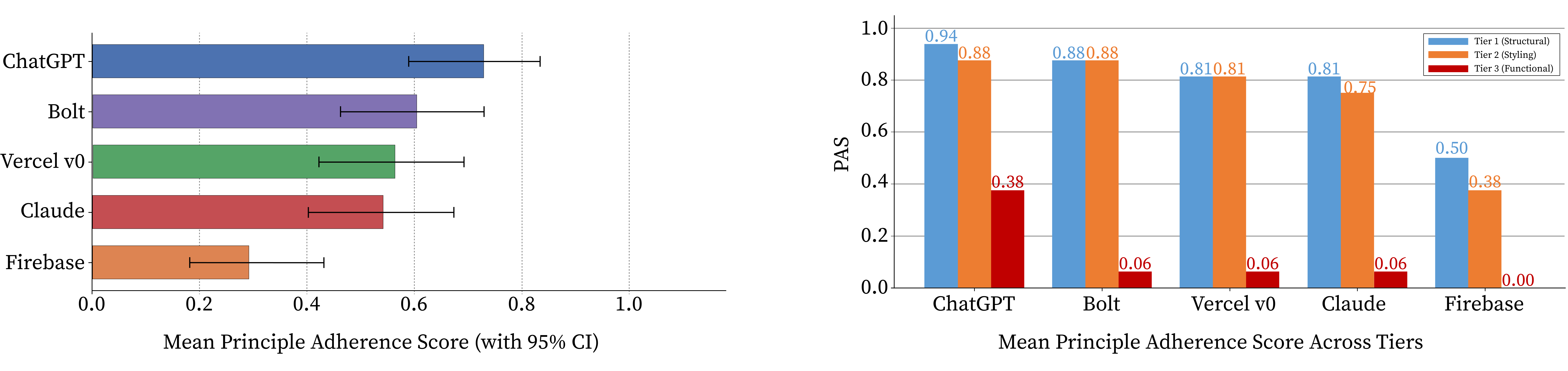}
\caption{Principle Adherence Score across five UI agent tools. Left: mean PAS with 95\% confidence intervals. Right: PAS by principle tier, showing consistently poor performance on functional principles}
\label{fig:PAS_graphs}
\end{figure*}
Our PAS metric measures whether generative UI tools implement the UX principles implicitly required by each design task. Figure \ref{fig:PAS_graphs} (left) reports the mean PAS for each tool across tasks, while Table \ref{tab:tool_performance} in its second column reports these same means with 95\% confidence intervals. These intervals capture uncertainty around each tool’s estimated average PAS across benchmark tasks. Overall, tools differed in how well they implemented the embedded UX principles. Mean PAS ranged from 0.29 for Firebase Studio to 0.73 for ChatGPT. In concrete terms, Firebase Studio implemented 14 of the 48 required principles, whereas ChatGPT implemented 35 of 48. Vercel v0 and Claude fell near the middle of the range, implementing only slightly more than half of the embedded UX principles. The non-overlapping 95\% bootstrap confidence intervals between the highest- and lowest-performing tools provide descriptive evidence that ChatGPT and Firebase Studio occupied meaningfully different parts of the PAS range.

Figure \ref{fig:PAS_graphs} (right) reports each tool’s mean PAS across the three task tiers. The clearest pattern is a sharp drop in PAS for functional tasks. Tools implemented structural tasks at relatively high rates, with Tier 1 scores ranging from 0.50 to 0.94, and styling tasks at similarly high rates, with Tier 2 scores ranging from 0.38 to 0.88. By contrast, functional tasks were implemented much less reliably. Four of the five tools scored 0.06 or below on Tier 3, and Firebase Studio implemented none of the functional principles across the eight functional tasks. Even ChatGPT, which scored highest overall, reached only 0.38 on Tier 3, indicating that functionality-related UX principles were the most difficult for these tools to implement. This pattern suggests that current tools share systematic weaknesses in interaction design principles such as visibility of system status, user control and freedom, and error prevention and recovery (i.e., they struggle implementing functional UX principles).

\begin{table}[h]
\centering
\resizebox{\columnwidth}{!}{%
\begin{tabular}{l|cccc}
\toprule
\textbf{Tool} & \textbf{Mean PAS [95\% CI]} & \textbf{Tier 1} & \textbf{Tier 2} & \textbf{Tier 3} \\
\hline
ChatGPT        & 0.73 [0.59, 0.83] & 0.94 & 0.88 & 0.38 \\
Bolt           & 0.60 [0.46, 0.73] & 0.88 & 0.88 & 0.06 \\
Vercel v0      & 0.56 [0.42, 0.69] & 0.81 & 0.81 & 0.06 \\
Claude         & 0.54 [0.40, 0.67] & 0.81 & 0.75 & 0.06 \\
Firebase       & 0.29 [0.18, 0.43] & 0.50 & 0.38 & 0.00 \\
\bottomrule
\end{tabular}%
}

\caption{Principle Adherence Score (PAS) comparison across generative UI tools.}
\label{tab:tool_performance}
\end{table}

 Table \ref{tab:tool_performance} reports the mean PAS score of each tool overall and separately for each task tier (columns 3–5). Because PAS measures whether a tool successfully implements the UX principles implicitly required by a task, higher PAS scores indicate that the generated interface more fully adheres to the expected UX principles. From Table \ref{tab:tool_performance}, we observe more variability between tools on structural (Tier 1) and styling (Tier 2) principles. ChatGPT and Bolt consistently achieved the highest PAS scores on these tiers. For example, ChatGPT achieved a mean PAS of 0.94 on structural principles and 0.88 on styling principles. In contrast, Firebase Studio achieved the lower score of 0.50 on structural principles and 0.38 on styling principles. This represents gaps of approximately 0.44 and 0.50 PAS points respectively relative to ChatGPT, indicating larger differences in the extent to which tools implemented these two requested UX principles. By contrast, the differences between tools on functional principles (Tier 3) were much smaller, with PAS scores ranging only from 0.00 to 0.38. However, this narrower range did not indicate similar performance quality across tools. Instead, it reflected a broader collapse in the ability of nearly all tools to implement functional UX principles at all, even when those principles were implicitly required by the task. Four of the five tools achieved Tier 3 PAS scores of only 0.06 or lower, while Firebase Studio failed to implement any functional principles across the Tier 3 tasks. 
 

\paragraph{Metric 3: Design Homogeneity Index (DHI)} Our DHI metric studies how similar the interfaces generated by the different tools are to one another, particularly when the tools receive the same prompt. We examined this similarity across three submetrics: visual similarity, color similarity, and layout similarity. 

\begin{figure*}[t]
\centering
\includegraphics[width=\linewidth]{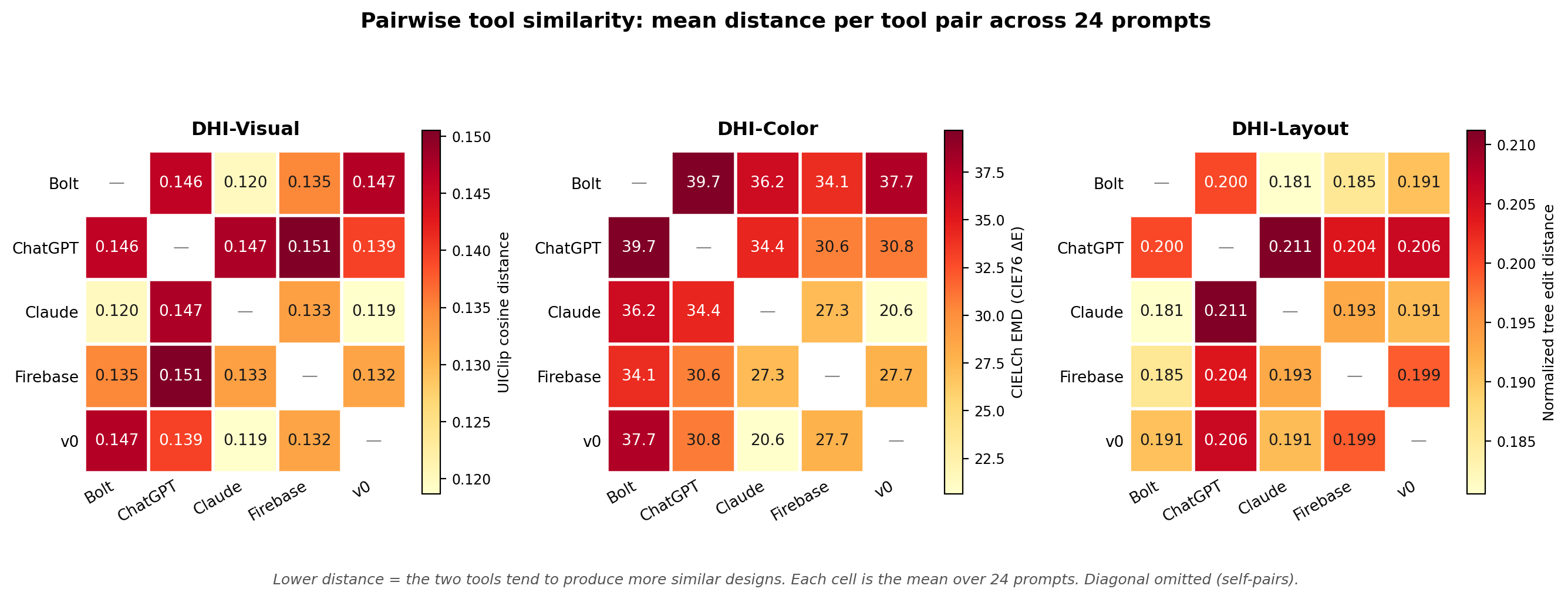}
\caption{Design Homogeneity Index Heat Map}
\label{fig:DHIheatmap}
\end{figure*}

Figure \ref{fig:DHIheatmap} visualizes the DHI between pairs of generative UI tools across the three DHI submetrics. Each cell represents the mean distance between the interfaces produced by two tools when given the same prompts, pooled across all three task tiers. Lower values indicate that two tools tended to generate more similar interfaces, while higher values indicate greater divergence in their generated designs. From the heatmap, we observe that similarity patterns varied across submetrics. For example, Claude and Vercel v0 produced the most visually similar interfaces (DHI-Visual = 0.119) and the most similar color palettes (DHI-Color = 20.6), while ChatGPT and Claude produced the most different layouts (DHI-Layout = 0.211). The heatmap highlights how some tools converged toward similar design patterns, whereas others generated interfaces with different visual, color, and layout characteristics.

Table \ref{tab:dhi_pair_summary} summarizes the most and least similar tool pairs for each DHI submetric. Each value represents the mean pairwise distance between the interfaces generated by two tools across the 24 tasks. Lower values indicate that two tools tended to generate more similar interfaces, while higher values indicate greater divergence in the generated designs. From Table \ref{tab:dhi_pair_summary}, we observe that Claude and Vercel v0 consistently produced some of the most similar interfaces, achieving the lowest distances on both visual similarity (0.119) and color similarity (20.6). In contrast, ChatGPT and Firebase produced the most visually different interfaces (0.151), while Bolt and ChatGPT produced the largest differences in color palettes (39.7). For layout similarity, Bolt and Claude generated the most similar layouts (0.181), whereas Claude and ChatGPT produced the most divergent layouts (0.211). The table also shows that the amount of variation differed across submetrics. DHI-Color exhibited the widest range of distances (20.6–39.7), indicating that overall tools varied in their color choices, while DHI-Visual and DHI-Layout showed narrower ranges, suggesting comparatively smaller differences in overall visual appearance and layout organization.

\begin{table}[t]
\centering
\normalsize
\setlength{\tabcolsep}{2pt}
\renewcommand{\arraystretch}{1.05}

\resizebox{\columnwidth}{!}{%
\begin{tabular}{l|l|l|c|c}
\toprule
\textbf{Sub-metric} & \textbf{Most similar pair} & \textbf{Least similar pair} & \textbf{Range} & \textbf{Mean} \\
\hline
DHI-Visual & Claude--v0 (0.119) & ChatGPT--Firebase (0.151) & 0.119--0.151 & 0.137 \\
DHI-Color  & Claude--v0 (20.6)  & Bolt--ChatGPT (39.7)      & 20.6--39.7   & 31.9 \\
DHI-Layout & Bolt--Claude (0.181) & Claude--ChatGPT (0.211) & 0.181--0.211 & 0.196 \\
\bottomrule
\end{tabular}%
}
\caption{Pairwise tool similarity by sub-metric, averaged across 24 tasks.}
\label{tab:dhi_pair_summary}
\end{table}

\section{Discussion}

\paragraph{Democratizing UI Generation Without Democratizing Evaluation}
Generative UI tools are often framed as democratizing design because they allow non-designers to rapidly prototype and generate functional interfaces from natural language \cite{takaffoli2024generative, chen2025genui, zhou2024exploring}. Yet this democratization also redistributes design labor. Activities that were once mediated by trained designers, such as prototyping, design justification, or  preliminary design evaluation, are increasingly being taken up by product managers, clients, and other non-design stakeholders. This shift creates new challenges around evaluation of the designs. Prior work shows that UX practitioners already use generative AI in design workflows \cite{takaffoli2024generative}, but also report a need for better training to assess the quality of the interfaces. For novice users, this problem is even more pronounced: they may be able to generate an interface, but lack the expertise to judge whether it reflects their goals, satisfies requirements, or follows sound UX principles. The challenge becomes more consequential when generative UI tools also provide user-facing explanations of what was designed and why \cite{sun2026seeing}. By articulating layout decisions, invoking design principles, and describing tradeoffs in the language of trained designers \cite{son2026clearfairy}, these systems can appear to perform design expertise on the user’s behalf. This may lead novice users to perceive generated interfaces as more complete, intentional, or correct than they actually are. In this way, generative UI tools can create a false sense of expertise: users may feel confident that an interface satisfies their design needs, but only because they lack the training to recognize usability problems, missing requirements, or flawed design assumptions.


Across our evaluation, roughly one in four stated design rationales did not fully appear in the generated interface. This gap became especially pronounced for functional tasks, where tools achieved a Tier 3 TFS of 0.66. The PAS results were even more striking. Recall that PAS measured whether generative UI tools implemented the UX principles implicitly required by each task. On functional UX principles, four of the five tools scored $\leq 0.06$, meaning they implemented almost none of the interaction-design requirements embedded in the tasks, such as visibility of system status, user control and freedom, error prevention, and recovery. These functional failures are also the hardest for non-experts to detect. A missing color choice or layout inconsistency is often visible in the rendered interface. By contrast, missing state management, inaccessible interactions, weak error recovery, or absent user control may not be obvious from surface-level inspection. Thus, the most consequential failures are often the least visible, while the non-expert designers now responsible for evaluating generated interfaces may be the least equipped to identify these errors \cite{takaffoli2024generative}.

Design evaluation depends on professional judgment developed through practice \cite{koskinen2013design}. This expertise is tacit, comparative, and built through repeated exposure to critique \cite{bardzell2014reading, haraway2013situated}. It does not transfer to non-experts simply because interface generation has become faster or more fluent. In fact, the ability to recognize when a design rationale sounds convincing but the interface does not behave correctly is precisely what may be lost when trained designers are removed from the loop \cite{takaffoli2024generative}. In this context, user-facing reasoning traces can begin to function like design documentation, but without the accountability structures that traditionally make such documentation meaningful. Prior work shows that users adjust their trust based on how rationales are presented, even when the underlying decision remains unchanged \cite{sun2026seeing, bertrand2022cognitive}. The central risk, then, is not simply that non-designers are using design tools. Rather, it is that the ease of generating interfaces may create a sense of competence in evaluating design decisions, even when users lack the expertise needed to distinguish sound design choices from hollow or problematic ones.

\paragraph{The Evaluation Bottleneck in Generative UI} 
Our findings point to a broader evaluation bottleneck in generative UI workflows. Generative UI tools make interface production faster, cheaper, and rhetorically polished, but evaluating the resulting artifact remains slow, expert-dependent, and multidimensional. A generated interface cannot be assessed only by whether it renders or whether the accompanying rationale sounds plausible. It must be evaluated as an artifact that combines design principles like visual hierarchy, interaction flow, accessibility, responsiveness, and contextual fit. Prior work on generative AI evaluation has warned that static and model-centric evaluations often miss modality, context, deployment conditions, and downstream harms \cite{rauh2024gaps}. In our study, this gap appears as a reasoning-to-artifact mismatch: generative UI tools can produce a professional-looking interface and a persuasive explanation in seconds, while the work of verifying whether the explanation is actually implemented remains externalized to users \cite{ibrahim2025towards}. Design Theater is therefore not only a failure of generation, but also a failure to produce artifacts that can be easily evaluated \cite{eriksson2025can}.

Our benchmark responds to this evaluation bottleneck by making three aspects of generated UI artifacts auditable. TFS evaluates whether an artifact implements the tool's design rationale. PAS evaluates whether an artifact satisfies UX principles implied by the prompt. DHI evaluates whether tools converge toward similar visual, color, and layout defaults despite presenting their outputs as situated design choices. Together, these metrics do not claim to measure interface quality in full. Instead, following prior work that emphasizes construct clarity in AI measurement \cite{bommasani2024trustworthy}, we treat TFS, PAS, and DHI as targeted instruments for detecting three failure modes: unimplemented design rationales, missed implicit UX principles, and homogenized design defaults. We therefore position the benchmark not as a final ranking of generative UI tools, but as a reusable evaluation framework for asking whether user-facing design reasoning is grounded in implementation.

\paragraph{Design Homogenization and the Loss of Situated Design}
Our DHI results suggest that Design Theater is not only a problem of missing implementation but also of reduced variation in generated UI artifacts. Prior work has shown that AI-assisted creation can narrow the diversity of results in writing and ideation \cite{dhruv2025WesternStyle, wadinambiarachchi2024effects}. In generative UI, homogenization is not only an output-level pattern; it is an accountability problem because similar artifacts arrive with rationales that frame them as context-sensitive design decisions \cite{ibrahim2025towards}. Interfaces are not only expressive artifacts; they organize attention, access, trust, navigation, and user action. This matters because design knowledge is situated: what counts as clear, trustworthy, accessible, or safe depends on the users, institutions, communities, and power relations a design is meant to serve \cite{costanza2020design}. When different users, domains, and stakes are repeatedly translated into similar visual and structural patterns, while being described as tailored design choices, generative systems risk flattening the contextual differences that design work is meant to preserve. This produces surface-level contextualization: the rationale names the audience, domain, or constraint, making the interface appear tailored, even when the rendered artifact relies on familiar defaults \cite{turpin2023language, sun2026seeing}. Convergence therefore carries stakes beyond aesthetics: when interfaces repeat the same patterns, they standardize where users look \cite{Jiang2023UEyes}, how they navigate and interact \cite{nielsen1994usability}, and who is included or excluded by design decisions \cite{costanza2020design}. When generated interfaces cluster visually or structurally while being presented as situated design responses, generative UI risks making repeated conventions appear unique and professionally validated.

Our DHI metric does not by itself determine whether a particular similarity is harmful, useful, or contextually appropriate. Rather, it makes visible when generated interfaces converge despite rationales that describe them as tailored to specific users and contexts. This allows us to ask whether AI-generated interfaces reflect meaningful adaptation to context or the repetition of dominant design defaults. Without design review, stakeholders may read a polished explanation as evidence that a generic interface has been meaningfully tailored \cite{kaur2020interpreting, sun2026seeing}.

\paragraph{Limitations and Future Work}
Our evaluation is artifact-centered. We inspect prompts, user-facing rationales, and rendered interfaces, but do not measure how designers, product managers, engineers, clients, or novice creators interpret these rationales. Our study therefore establishes that rationale-to-artifact mismatch exists, but not how often stakeholders detect it or how it affects trust, review, and deployment decisions. We evaluate Design Theater across five generative UI tools, enabling systematic comparison across tools, tiers, and metrics, but this scope should not be interpreted as an exhaustive account of generative UI behavior. These systems evolve rapidly, and outputs may vary across model versions, interface defaults, and repeated generations.
Our benchmark required tools to generate interfaces using only HTML, CSS, and JavaScript, improving comparability while excluding component libraries used by some tools. Because tools were evaluated only against commitments they chose to make, this constraint does not explain TFS results. It may, however, affect PAS, where implementing functional principles without libraries requires additional effort, and DHI, where shared constraints may reduce output variation. Our metrics capture specific dimensions of Design Theater rather than overall design quality. TFS measures whether rationales are implemented, PAS measures whether selected implicit UX principles are recognized and implemented, and DHI measures visual, color, and layout convergence across tools. DHI also lacks a human-designed reference distribution, meaning our results describe convergence among the evaluated tools rather than deviation from human practice. These metrics do not capture aesthetic judgment, cultural fit, emotional resonance, accessibility in lived use, or user success in deployed contexts. Future work should extend this evaluation to additional tools, repeated samples, multi-screen flows, mobile interfaces, multi-turn workflows, accessibility audits, usability testing, participatory evaluation, and multilingual or culturally situated scenarios.

\section{Conclusion}
Generative UI tools produce user-facing design rationales alongside generated interfaces, but whether these rationales reflect implementations remains unexamined. We introduce Design Theater to characterize this gap and propose a benchmark with three metrics: Thinking Fidelity Score, Principle Adherence Score, and Design Homogeneity Index. Across designs from five state-of-the-art tools, we identify three patterns. First, roughly one in four design rationales fails to appear in the generated interface. Second, tools recognize only about half of the UX principles required by their tasks, with near-universal failures on functional principles such as visibility of system status, user control and freedom, and error prevention and recovery. Third, generated interfaces converge across tools, showing narrow differences in visual appearance and layout organization, while color choices vary more widely. As reasoning traces become a primary signal of design competence for non-expert reviewers, measuring the gap between what tools say and what they build is critical for responsible deployment of generative UI tools.

\section*{Acknowledgments}
Special thanks to the anonymous reviewers for their feedback. This work was partially supported by NSF grants 2339443 and 2403252. We used Claude Sonnet 4.6 and Grammarly to refine language and clarity under full author oversight; all study design, analysis, literature review, and writing were conducted and verified by the authors.

\bibliography{aaai2026}

\appendix

\section{Appendix}

\section{Data and Materials Availability}

The full evaluation corpus, including the design task prompts,
generated UI artifacts, user-facing reasoning traces, and outputs
produced by each tool, is publicly available at:

\begin{center}
\url{https://github.com/kashifimteyaza/design-theater.io}
\end{center}

The underlying models used by the tools during the study included GPT-5 Thinking (ChatGPT), Claude Sonnet 4.5 (Claude and Bolt), Claude Haiku 4.5 (Vercel v0), and Gemini 2.5 Pro (Firebase Studio).

\end{document}